\documentclass[conference]{IEEEtran}
\IEEEoverridecommandlockouts
\usepackage{cite}
\usepackage{amsmath,amssymb,amsfonts}
\usepackage{subcaption}
\usepackage{algorithmic}
\usepackage{algorithm}
\usepackage{graphicx}
\usepackage{textcomp}
\usepackage{xcolor}
\usepackage{multirow}
\usepackage[normalem]{ulem}
\useunder{ine}{}{}
\usepackage{amsthm}
\theoremstyle{theorem}

\theoremstyle{definition}

\def\BibTeX{{\rm B\kern-.05em{\sc i\kern-.025em b}\kern-.08em
    T\kern-.1667em\lower.7ex\hbox{E}\kern-.125emX}}

\makeatletter
\newcommand{\linebreakand}{%
  \end{@IEEEauthorhalign}
  \hfill\mbox{}\par
  \mbox{}\hfill\begin{@IEEEauthorhalign}
}
\makeatother

\begin{document}

\title{Federated Knowledge Graph Completion via Latent Embedding Sharing and Tensor Factorization\\
}

\author{
\IEEEauthorblockN{ Maolin Wang}
\IEEEauthorblockA{
\textit{City University of Hong Kong} \\
Hong Kong, China \\
morin.w98@gmail.com}
\and
\IEEEauthorblockN{Dun Zeng}
\IEEEauthorblockA{
\textit{UESTC}\\
Chengdu, China \\
zengdun@std.uestc.edu.cn}
\and
\IEEEauthorblockN{ Zenglin Xu}
\IEEEauthorblockA{
\textit{Harbin Institute of Technology~(Shenzhen)}\\
Shenzhen, China \\
zenglin@gmail.com}
\linebreakand
\IEEEauthorblockN{ Ruocheng Guo}
\IEEEauthorblockA{
\textit{ByteDance Research}\\
London, UK \\
rguo.asu@gmail.com}
\and
\IEEEauthorblockN{Xiangyu Zhao}
\IEEEauthorblockA{\IEEEauthorblockA{
\textit{City University of Hong Kong} \\
Hong Kong, China \\
xianzhao@cityu.edu.hk}}}
\maketitle

\begin{abstract}

Knowledge graphs (KGs), which consist of triples, are inherently incomplete and always require completion procedure to predict missing triples. In real-world scenarios, KGs are distributed across clients, complicating completion tasks due to privacy restrictions. Many frameworks have been proposed to address the issue of federated knowledge graph completion. However, the existing frameworks, including FedE, FedR, and FEKG, have certain limitations. = FedE poses a risk of information leakage, FedR's optimization efficacy diminishes when there is minimal overlap among relations, and FKGE suffers from computational costs and mode collapse issues. To address these issues, we propose a novel method, i.e., Federated Latent Embedding Sharing Tensor factorization (FLEST), which is a novel approach using federated tensor factorization for KG completion. FLEST decompose the embedding matrix and enables sharing of latent dictionary embeddings to lower privacy risks. Empirical results demonstrate FLEST's effectiveness and efficiency, offering a balanced solution between performance and privacy. FLEST expands the application of federated tensor factorization in KG completion tasks.
\end{abstract}


\section{Introduction}

In the field of knowledge representation and reasoning, a knowledge graph (KG) is a knowledge base that integrates data using a graph-structured data model or topology~\cite{guo2020survey}. A KG displays the relationship between a network of real-world elements, such as objects, events, circumstances, or concepts.
{KGs are widely used in various highly influential applications, like recommendation systems~\cite{guo2020survey}.}
Collected large-scale KGs, like Freebase~\cite{bollacker2008freebase}, always contain millions of entities and relationships. However, the huge collection cost and observation bias of the real world will lead to the incomplete content of the KGs. Many missing facts and implicit relationships need to be fully uncovered. The missing facts are represented as the 0 entries in the tensor. It is necessary to determine which of these 0 values are corresponding to missing facts, which impedes us to fully leverage the information embedded in the KG.
This collection of issues is known as the KG completion problem or the link prediction problem~\cite{rossi2021knowledge}.
To tackle the problem of incomplete KGs, researchers have been exploring approaches to represent relations and entities as continuous vectors. This enables mapping knowledge graph problems into mathematical optimization problems in vector spaces. A series of popular KG embedding methods have achieved great success, including DistMult~\cite{yang2015embedding}, ComplEx~\cite{trouillon2016complex}, and RotatE~\cite{sun2018rotate}.
%

These existing methods all require to store an entire KG in one device. However, in some real-world situations, KGs are distributed across different organizations (e.g., companies or hospitals)~\cite{huang2022fedcke}. Due to the need for user private information protection, the knowledge collected by different organizations cannot be shared. 
Utilizing the complementary capabilities of various distributed but related KGs while maintaining such protection of privacy poses a pressing challenge in real-world KG applications. Fortunately, such issue can be solved through the paradigm of federated learning~(FL)~\cite{JMLR:v24:22-0440,chen2021fede}. FL enables different clients to collaborate in learning global knowledge without sharing their local data~\cite{ JMLR:v24:22-0440}.

One key technique to improve the performance
of KG embedding methods in a federated learning paradigm is to align the embeddings of overlapping
entities across KGs. FedE~\cite{chen2021fede} is the initial solution that presents a mechanism in which the server maintains an extensive table consisting of entity embeddings and corresponding entity IDs. \textbf{This allows the server to infer relationships easily and will result in a high private information leakage risk.} Later, FedR was introduced to address privacy concerns, specifically concentrating on aligning relation embeddings. \textbf{However, 
when there is little overlap among relations across clients, the effectiveness of FedR~\cite{zhang2022efficient} can be significantly degraded.}
Unlike server-side alignment, FKGE~\cite{peng2021differentially} facilitates entity alignment between clients. Taking inspiration from PATEGAN~\cite{jordon2018pate}, FKGE incorporates an adversarial translation (PPAT) network for adversarial learning. \textbf{However, it is important to note that this approach may suffer from huge computational costs and mode collapse issues inherent in adversarial training~\cite{huang2023mitigating,creswell2018generative}. These factors can impact the robustness of the training process.}

\begin{figure}[t]
\centering
 \includegraphics[width=0.45\textwidth]{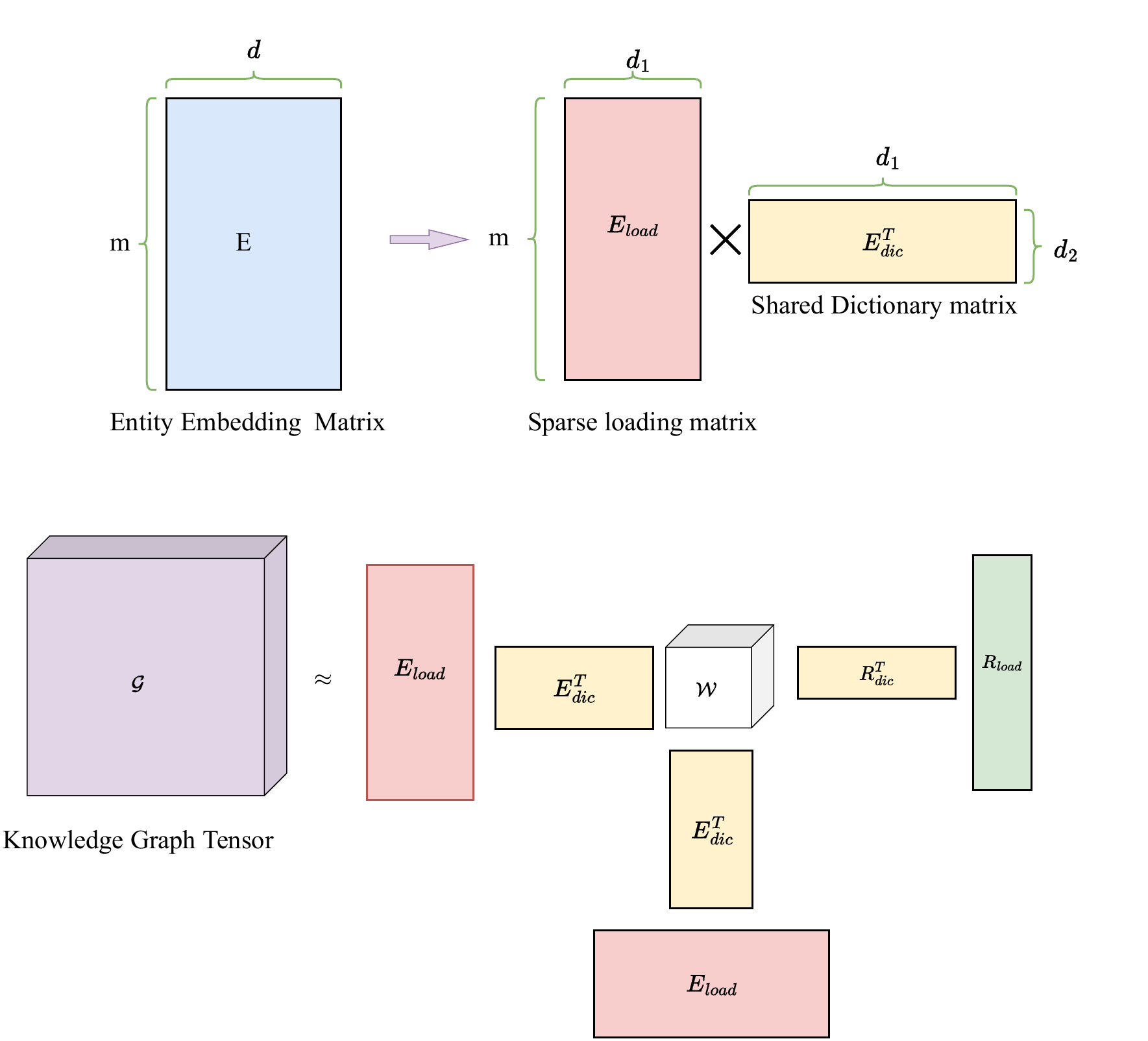}
  \caption{
  {In order to have a lower private information leakage risk in each mode in federated tensor decomposition scenarios, we can divide the embedding matrix into a dictionary matrix and a loading matrix.} In this way, the adjacency tensor of the entire KG can be decomposed into the form in this figure. We can share the latent dictionary embedding matrix. Even if the latent dictionary embedding matrix is leaked, the specific entity level and relationship level information cannot be recovered.
  }
  \label{fig:dictionary}
 \end{figure}

Given the successful combination of federated learning and tensor factorization~\cite{kim2017federated,ma2021communication,li2022personalized}, as well as the probabilistic tensor representation of knowledge graphs, we aim to propose a federated tensor factorization framework for knowledge graph completion to overcome the limitations of these existing models. However, most of the existing federated tensor factorization methods, such as TRIP~\cite{kim2017federated}, share the information of certain tensor modes and preserve other tensor/matrix modes, which makes it unsuitable for federated knowledge graphs problems.
Therefore, we propose a novel simple and effective framework, \textbf{F}ederated \textbf{L}atent \textbf{E}mbedding \textbf{S}haring {and} \textbf{T}ensor factorization~(FLEST), to address this issue. Our basic idea is depicted in Fig~\ref{fig:dictionary}.
The FLEST model decomposes the embedding matrix into a dictionary matrix and a loading matrix. The adjacency tensor of the entire KG can also be decomposed via a Tucker-like format in this manner. By sharing the latent dictionary embedding matrix within each mode, even if the dictionary matrix is leaked and the number of latent dimensions is much smaller than the original number of entities, it remains infeasible to recover specific entity-level and relationship-level information.

\section{Methods}
In this section, we will introduce our methods,  which is named as Federated Latent Embedding Sharing Tensor factorization scheme~(shorten as FLEST)

\subsection{Tensor Preliminaries}

{Tensor contraction~\cite{balazevic2019tucker,shi2022cardinality} means that two tensors are contracted into one tensor along their associated pairs of indices. Given two tensors ${ \mathcal{A}} \in \mathbb{R}^{I_1\times I_2\times \cdots \times I_N}$ and ${\mathcal{B}}\in\mathbb{R}^{J_1\times J_2\times  \cdots \times J_M}$, with some common modes, $I_{n_1}=J_{m_1}$, $\cdots$ $I_{n_S}=J_{m_S}$, the tensor contraction ${\mathcal{A}} \times^{(n_1,n_2\cdots,n_S)}_{(m_1,m_2\cdots,m_S) } {\mathcal{B}}$ yields a $(N+M-2S)$-order tensor ${\mathcal{C}}$.
Tensor contraction can be formulated as:
\begin{align}
     &\mathcal{C} = \mathcal{A} \times_{(j_{m_1}, j_{m_2}, \dots j_{m_S})}^{(i_{n1}, i_{n2}, \dots i_{n_S})} \mathcal{B}  \notag\\
    &= \sum_{i_{1},i_{2},\cdots i_{N}} \mathcal{A}_{i_1,i_2,\cdots i_{n_S},*}\quad\mathcal{B}_{*,i_1,i_2,\cdots i_{n_S}}.
\end{align}

The well-known Mode-$N$ Product is a special case of Tensor Contraction. Given a tensor ${ \mathcal{A}} \in \mathbb{R}^{I_1\times I_2\times \cdots \times I_N}$ and a matrix ${\mathbf{B}}\in\mathbb{R}^{J_1\times J_2}$. If $J_2 = I_{n}$, then
\begin{equation}
    {\mathcal{C}}
    =  {\mathcal{A}} \times^{(n)}_{(2) } {\mathbf{B}} = {\mathcal{A}} \times_{n} {\mathbf{B}}.
\label{eq:tensor_contraction}
\end{equation}

\subsection{Preliminaries for Graph and Knowledge Graph}

A graph~\cite{fu2022federated} is represented by the formula $G = (V, E),$ where $V$ is the set of vertices and $E$ is the set of paired vertices, or edges. A graph may be represented by its adjacency matrix $\mathbf{A}\in \{0,1\}^{|V|\times |V|}$, where $|\cdot|$ represents the cardinality of the set. The elements of $\mathbf{A}$ indicate whether or not two vertices in the graph are adjacent.  $\mathbf{A}_{i,j} = 1$ if $\left\{v_i, v_{j}\right\} \in {E}$ and $0$ otherwise. 
A knowledge graph~\cite{balazevic2019tucker} characterizes the ordered triplets $(e_s, r, e_o)$ of an entity set $E$ and a relationship set $R$. A knowledge graph $G = (E,R)$ can be represented as a third-order adjacency tensor $\mathcal{A} \in\{0,1\}^{|E| \times |R| \times |E|}$. 
\subsection{Problem and Model Formulation}
Based on the tensor representation of the knowledge graph and the Tucker format~\cite{balazevic2019tucker} 
, we can formulate the knowledge graph completion problem in a probability tensor decomposition framework.
Given the knowledge graph $G = (E,R)$, we let $\mathcal{A}={a_{irj}}\in(0,1)^{|E| \times |R| \times |E|}$, with $a_{irj} = 1$ when the triplet $(e_i,r,e_j)$ exists in $G$ and $a_{irj}=0$ when it does not exist. Let $\Theta = (\theta_{irj})$ be the {entry-wise} transformation of $\mathcal{A}$ is shown as:

\begin{align}
   \theta_{irj} = log\left(\frac{a_{irj}}{s-a_{irj}}\right) ,
   \label{equation1}
\end{align}

where $s$ is the sparsity factor and is introduced to describe the sparsity of the original KGs
. The modified logit transformation of Eq.~\ref{equation1} implies that
$
a_{irj} = s\left(1+e^{-\theta_{irj}}\right)^{-1}
$. 
Thus, we can consider applying tensor decomposition to a continuous tensor $\Theta$ to model the original binary tensor $\mathcal{A}$.
We consider adopting the following tensorial knowledge graph embedding model in the Tucker format~\cite{bi2022tensor},
\begin{align}
\Theta \approx \hat{\Theta} = \mathcal{W}\times_1E\times_2 R\times_3 E,
   \label{equation2}
\end{align}
where $E\in \mathbb{R}^{|E|\times r}$ is the entity embedding matrix,  
$R \in \mathbb{R}^{|R|\times r}$ is the relationship embedding matrix, and $\mathcal{W}\mathbb{R}^{r\times r\times r}$ is the non-symmetric third-order core tensor. $r$ is the number of ranks. In this paper, all ranks in the tensor decomposition are assumed to have the same value for simplicity.
Given the knowledge graph tensor $\mathcal{A}$ and the  embedding model in Eq.~\ref{equation2}, 
the negative log-likelihood loss is: 
\begin{align}
&\mathcal{L}_{ll}\left(\hat{\Theta} ; \mathcal{A}\right)= - \sum_{i,r,j}\log \left(1+\frac{s}{1-s+e^{-\hat{\theta}_{irj}}}\right)-\notag\\
&a_{irj} \log \left(\frac{s}{1-s+e^{-\hat{\theta}_{irj}}}\right).
\label{eq:likeli}
\end{align}

This formulation provides a solid probabilistic tensor decomposition model.
However, it is still  extremely challenging to generalize it to the setting of federated tensor decomposition.
This is because, in the previously mentioned scenario of tensor decomposition, such as TRIP~\cite{kim2017federated} and FGTF~\cite{ma2021communication}, some safe dimension information could be shared. For example, in the case of triplets of healthcare~\cite{kim2017federated}, (patients, medication, diagnosis), only patients' embedding information needs to be protected, and we can share medication embedding and diagnosis embeddings.
However, in the context of a knowledge graph~\cite{zhang2022efficient,peng2021differentially}, it becomes necessary to protect both entities and relationships. To address a full mode of private information protection, we propose to adopt a strategy of sharing hidden variables. 
As illustrated in Fig.~\ref{fig:dictionary}, we partition the original embedding matrix into two matrices and perform a multiplication operation to reconstruct the original matrix.

The matrix of latent variables is referred to as the dictionary matrix, which can be viewed as a representative encoding of the original data. The other matrix, the loading matrix, represents the linear combination weights to load representative features.
And if we only share the latent dictionary matrix at this point, it is almost impossible to infer any entity or triplet from relation or latent embedding only. 

Formally, as shown in Fig.~\ref{fig:dictionary}, we can express $ \Theta $ as
\begin{align}
\Theta =\mathcal{W}\times_1 E_{dic}E_{loading}\times_2 R_{dic}R_{loading}\times_3 E_{dic}E_{loading},
\end{align}
where $\mathcal{W}\in\mathbb{R}^{r\times r\times r}$ is the fusion weight tensor, $E_{dic}\in\mathbb{R}^{r\times r}$ is the entity dictionary matrix, $R_{dic}\in\mathbb{R}^{r\times r} $is the relationship dictionary matrix, $E_{loading}\in\mathbb{R}^{r\times |E|} $ is the entity loading matrix, $R_{loading}\in\mathbb{R}^{r\times |R|} $ is the relation loading matrix and $r$ is the number of rank.

Due to the fact that the core tensor often has massive parameters and tensor multiplication is extremely computationally expensive, we can use CP decomposition~\cite{bi2022tensor} to decompose the original fusion weight tensor $\mathcal{W}$. 
And if there are multiple clients, then the theta of a specific client $c$ can be represented:
\begin{align}
&\Theta^{(c)} =\mathcal{I}\times_1 W_{1} E_{dic}E^{(c)}_{loading}\times_2 W_{2} R_{dic}R^{(c)}_{loading } \notag\\ \times_3  
& W_3 E_{dic}E^{(c)}_{loading}.
\end{align}


where ${W}_{1}\in\mathbb{R}^{r\times r}$, ${W}_{2}\in\mathbb{R}^{r\times r}$, and ${W}_{3}\in\mathbb{R}^{r\times r}$ are the decomposed fusion weights and $\mathcal{I}$ is the identity tensor.

Furthermore, we assume that the vectors in the dictionary should exhibit a high degree of left orthogonality. (Here, we define the matrix $E$ as left orthogonal if $E^TE = I$.) This requirement is motivated by the need for distinctiveness among hidden features, similar to the significant differences observed in the specific functions of different gene expressions. Additionally, we aim for sparsity in the load matrix, considering that instances should possess several distinctive features.

Therefore, in addition to the likelihood error of Eq.~\ref{eq:likeli}, we also added sparse constraints to the loading matrix and orthogonal constraints to the dictionary matrix. The specific implementation is as follows:
\begin{align}
 &\mathcal{L}_{dic} = ||E^{T}_{dic}E_{dic} - I||_{F}+||R^{T}_{dic}R_{dic} - I||_{F}
 \label{eq:dic}
\\
&\mathcal{L}^{(i)}_{loading} = ||vec(E^{(i)}_{loading})||_{1}+||vec(R^{(i)}_{loading})||_{1},\label{eq:loading}
\end{align}
where $||\cdot||_F$ is the Frobenius norm.

So the final loss function could be represented as:
\begin{equation}
    \mathcal{L} = \sum_{i=1}^{C}\mathcal{L}_{ll}\left(\hat{\Theta}^{(i)} ; \mathcal{A}^{(i)}\right) + \alpha  \mathcal{L}_{dic} + \beta \mathcal{L}^{(i)},
    \label{eq:final_loss}
\end{equation}
where $\alpha$ and $\beta$ are non-negative hyperparameters controlling the trade-off among the two penalty terms.
\begin{figure}[t]
\centering
 \includegraphics[width=0.45\textwidth]{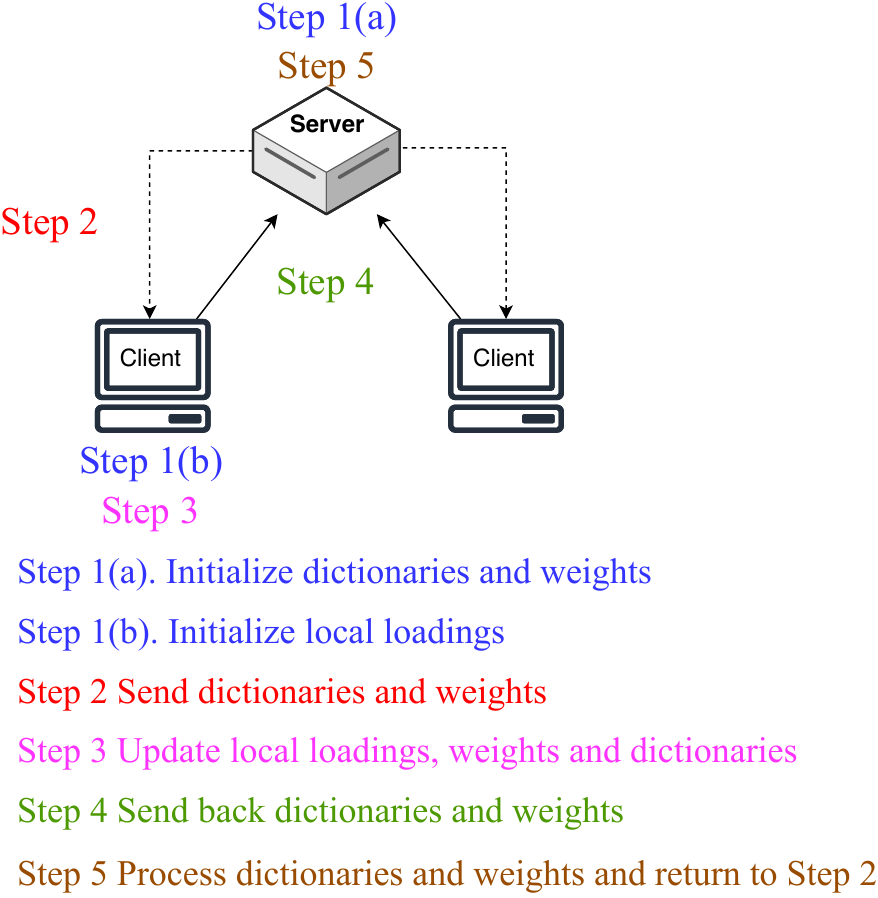}
  \caption{
  Illustration of the algorithm of FLEST, which consists of five steps. First, all clients and the central server initialize all parameters. Second, all clients start to accept the latent embedding dictionary from the central server. Subsequently, all parameters are updated according to the local data. After several epochs of training, all clients send parameters back to the server. The server averages the parameters and returns the distribution to all clients. Finally, after a few rounds of training, the training can be terminated. }
  \label{fig:dictionary3}
 \end{figure}

\subsection{Federated Optimization}
We can easily find the optimization in terms of $E_{dic}$ and $R_{dic}$ can be regarded as an example of the generalized FL optimization scheme. The sharing and protected parameters are shown in Fig~\ref{fig:dictionary3}.
The optimization problem in Eq.~\ref{eq:final_loss} can be solved via gradient-based methods in practice as computing the gradient for each parameter is straightforward.
We define $\mathcal{T}^{(c)} = 
\frac{\partial\mathcal{L}_{likelihood}\left(\hat{\Theta}^{(i)}\right)}{\partial\hat{\Theta}^{(i)}}
$.
As for $W_1$, $W_2$, $W_3$, we take $W_1$ as a example:
\begin{align}
& \nabla_{W_1} \mathcal{L}\left(W_1, W_2, W_3 ,E_{dic},E_{loading}^{(1)},\cdots\right)=\sum_{i=1}^{C} \notag \\ 
   &
   \mathcal{T}^{(i)} \times_{2,3}^{2,3}
(\mathcal{I}\times_2 W_2 R_{dic}R^{(i)}_{loading}\times_3 W_3 E_{dic}E^{(i)}_{loading}) 
\notag\\
&\times_1 E_{loading}^{(i)T}E_{dic}^T .
\label{W_1}
\end{align}
As for the dictionary matrix and loading matrix, we can get the full gradient if all triplets are given:
\begin{align}
&\nabla_{E_{dic}}\mathcal{L}\left(W_1,W_2,W_3,E_{dic},E^{(i)}_{loading},\cdots\right)
\notag\\
&= \sum_{i=1}^{C} \Delta_{(E_{dic})}^{(i)} = \sum_{i=1}^{C}
\mathcal{T}^{(i)} \times_{2,3}^{2,3}\notag\\
&
(\mathcal{I}\times_1 W_1\times_2 W_2 R_{dic}R_{loading}^{(i)}\times_3 W_3 E_{dic}E_{loading}^{(i)}) 
\notag\\
&\times_1 E_{loading}^{(i)T}
+\mathcal{T} \times_{1,2}^{1,2}
(\mathcal{I}\times_1 W_1 E_{dic}E_{loading}^{(i)}\times_2 \notag\\
&
W_2 R_{dic}R_{loading}^{(i)}
\times_3 W_3 ) \times_1 E_{loading}^{(i)T} + \notag\\
&4\alpha(E_{dic}E_{dic}^TE_{dic}-E_{dic}) ,
\label{E_{dic}}
\end{align}

\begin{align}
   & \nabla_{E_{loading}^{(i)}} \mathcal{L}^{(t)}\left(W_1,W_2,W_3,E_{dic},E_{loading}\right)
   =  \Delta_{E_{loading}^{(i)}}^{(i)}
   \notag
   \\&=\mathcal{T}^{(i)} \times_{2,3}^{2,3}
(\mathcal{I}\times_1 W_1E_{dic}\times_2 W_2 R_{dic}R_{loading}^{(i)}\times_3 \notag
\\&W_3 E_{dic}E_{loading}^{(i)})+
\mathcal{T} \times_{1,2}^{1,2}
(\mathcal{I}\times_1 W_1E_{dic}E_{loading}\times_2\notag\\& W_2 R_{dic}R_{loading}\times_3 W_3 E_{dic})+ \beta sgn(E_{loading}^{(i)}).
\label{E_{loading}}
\end{align}

We can find that within the framework of FL~\cite{mcmahan2017communication}, each client can calculate updates for parameters locally based on their own data and communicate via sharing dictionary matrices.
We can adopt FedAvg (averaging the parameter)~\cite{mcmahan2017communication} to design federated algorithms. Each client optimizes $E_{dic}$, $R_{dic}$, $W_1$, $W_2$ and $W_3$ locally using gradient-based methods (such as SGD). For example, for clients $i$, we can update local dictionary in the $n$-th column of $E_{dic}$ via
$$
E_{dic}^{(i)(n)} = E_{dic}^{(i)(t)} - \eta \Delta_{(E_{dic})}^{*(i)},
$$

where $\Delta_{(E_{dic})}^{*(i)}$ are gradient from a given mini-batch.
Then the same update applies to local parameters,
After multiple epochs, dictionaries  $E_{dic}, R_{dic}$ and the fusion weights $W_1,W_2$ and $W_3$ are uploaded to the server, and all parameters are averaged and distributed once by the server.
For example, the global $E_{dic}$ can be obtained as
$$
E_{dic} = \frac{1}{|C|}\sum_{i}^{C} E_{dic}^{(i)(N)}.
$$
The whole algorithm is represented in Fig.~\ref{fig:dictionary3}.

Compared with entity aggregation in FedE~\cite{chen2021fede}, it is almost impossible to infer any entity or triplet from a relation or latent dictionary embedding matrix leakage in the framework of FLEST. Different from FLEST,
FedE~\cite{chen2021fede} employs a mechanism where the server maintains a comprehensive table comprising entity embeddings and their corresponding entity IDs. This setup, as illustrated in Fig~\ref{fig:FedE}, facilitates the server's ability to infer relationships. It is important to note that this approach also introduces a significant risk of private information leakage. Compared with relationship aggregation in FedR~\cite{zhang2022efficient}, when there is little to no overlap among relations across clients, FLEST could still communicate effectively while FedR will reduce into a local-only scheme. It is also important to note that our proposed FLEST will not suffer from huge computational costs and mode collapse issues inherent in adversarial training in FKGE~\cite{peng2021differentially}.

\begin{figure}[t]
\centering
 \includegraphics[width=0.5\textwidth]{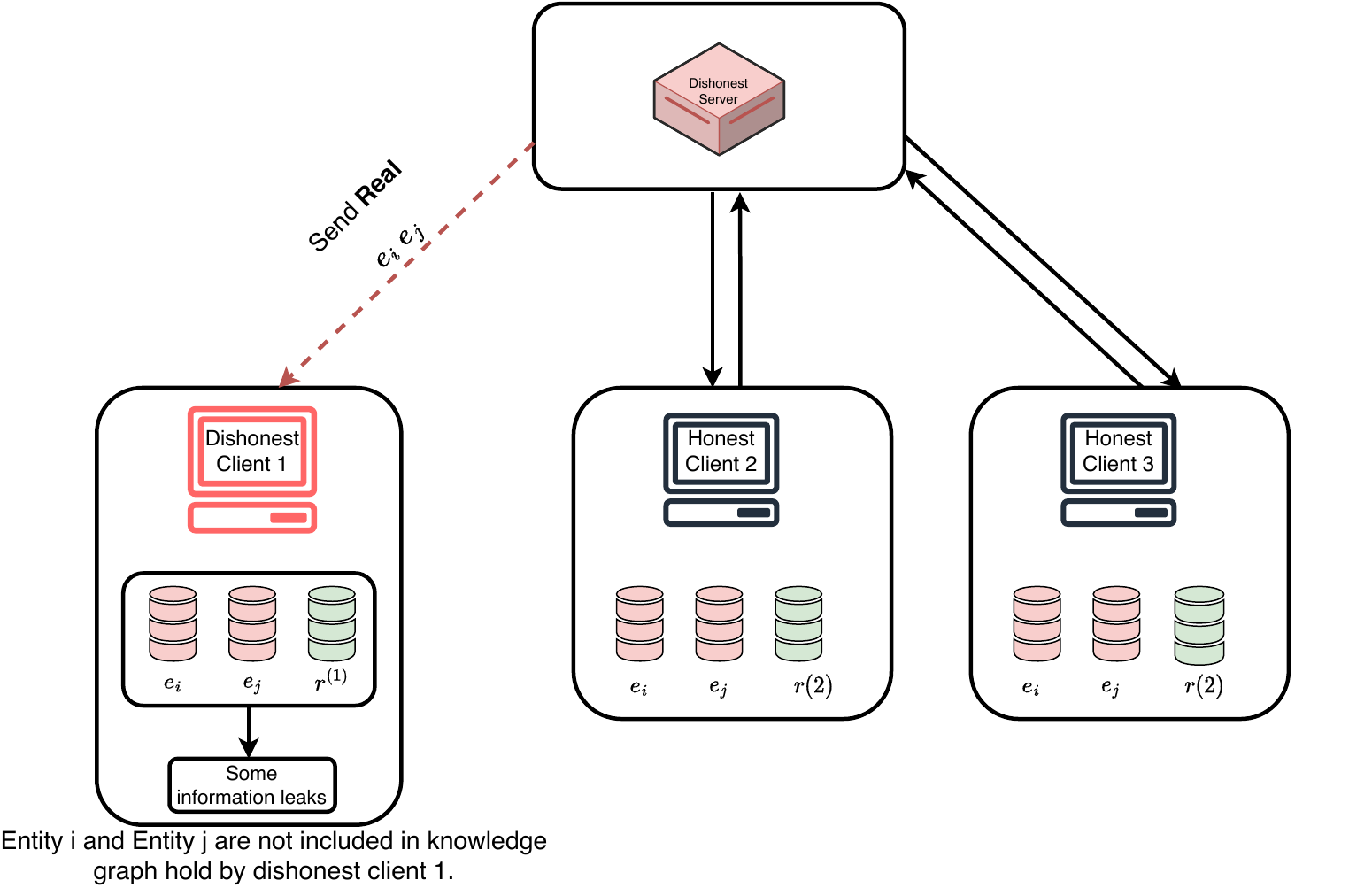}
  \caption{This is an attack on FedE~\cite{chen2021fede}. The dishonest client 1 and server collude. The entities i and j that do not exist in client 1 are leaked to client 1, and the relationship between $e_i$ and $e_j$ may be deduced accordingly. And $e_i$ and $e_j$ do not belong to client 1, but are unique users of client 2 and client 3, but their triple relationship has the risk of leakage~\cite{zhang2022efficient}.}
  \label{fig:FedE}
 \end{figure}

\section{Experiments}

\subsection{Dataset}
We evaluate our model using two standard knowledge graph datasets for link prediction:
\textbf{FB15k-237}~\cite{toutanova2015representing} was obtained by removing the inverse of multiple relationships present in the training set from both the validation and test sets of FB15k~\cite{toutanova2015representing}.
\textbf{WN18RR}~\cite{dettmers2018convolutional} is a link prediction dataset that has been derived from WN18, a subset of WordNet~\cite{dettmers2018convolutional}. 

\subsection{Evaluation Metrics}

We employ evaluation metrics commonly used in the link prediction literature~\cite{chen2021fede,zhang2022efficient,huang2022fedcke}:
\paragraph{MRR} Mean Reciprocal Rank (MRR) is a metric commonly used in information retrieval and recommendation systems. It measures the quality of the $i$-th predicted ranking by its reciprocal rank -- the inverse of the ranking of the highest-ranked correct answer $\frac{1}{\text{rank}_i}$. 
\paragraph{Hit@k} Hit@k metrics, such as Hit@10, Hit@3, and Hit@1, are widely used in evaluating top-k recommendation or information retrieval systems. They measure the proportion of predicted rankings that rank at least one correct answer among the top-k. For example, Hit@10 is the percentage of predicted rankings where at least one correct answer is ranked among the top-10 positions. 
These metrics provide valuable insights in ranking-based tasks, allowing us to assess whether a ranking model can accurately retrieve relevant information.
\begin{table}[htbp]
\caption{Scoring functions of baseline models}
  \centering
  \begin{tabular}{lc}
    \hline
    Model & Scoring Function \\
    \hline
    DistMult~\cite{yang2015embedding} & $\langle \mathbf{e}_s, \mathbf{r}, \mathbf{e}_o \rangle$ \\
    ComplEx~\cite{trouillon2016complex} & $\operatorname{Re}(\langle \mathbf{e}_s, \mathbf{r}, \overline{\mathbf{e}}_o \rangle)$ \\
    RotatE~\cite{sun2018rotate} & $-\| \mathbf{h} \circ \mathbf{r} - \mathbf{t} \|$ \\
    TuckER~\cite{balazevic2019tucker} & $\mathcal{W} \times_1 \mathbf{e}_s \times_2 \mathbf{r} \times_3 \mathbf{e}_o$ \\
    \hline
    Ours & $\mathcal{I} \times_1 W_1E_{dic} \mathbf{e}_s \times_2 W_1R_{dic} \mathbf{e}_s \mathbf{r} \times_3 W_1E_{dic} \mathbf{e}_s \mathbf{e}_o$ \\
    \hline
  \end{tabular}
  \label{tab1my}
\end{table}

\begin{table}[!htpb]
\caption{Results of Single Client Performance}
\centering
\resizebox{\linewidth}{!}{ 
\begin{tabular}{|l|llll|llll|}
\hline
          & \multicolumn{4}{c|}{WN18RR}   & \multicolumn{4}{c|}{FB15k-237} \\ \hline
\textbf{Metric} &
  \multicolumn{1}{l|}{MRR} &
  \multicolumn{1}{l|}{{ Hit@10}} &
  \multicolumn{1}{l|}{{ Hit@3}} &
  { Hit@1} &
  \multicolumn{1}{l|}{{ Hit@10}} &
  \multicolumn{1}{l|}{{ Hit@3}} &
  \multicolumn{1}{l|}{{ Hit@1}} &
  MRR \\ \hline
DistMult  & 0.431 & 0.490 & 0.451 & 0.393 & 0.421  & 0.266 & 0.165 & 0.254 \\ \hline
ComplEx   & 0.440 & 0.510 & 0.460 & 0.410 & 0.428  & 0.275 & 0.158 & 0.247 \\ \hline
HypER     & 0.435 & 0.522 & 0.477 & 0.436 & 0.524   & 0.376 & 0.252 & 0.341 \\ \hline
RotatE    & 0.476 & 0.571 & 0.492 & 0.428 & 0.533  & 0.375 & 0.241 & 0.338 \\ \hline
TuckER    & 0.473 & 0.546 & 0.482 & 0.443 & 0.540  & 0.394 & 0.266 & 0.358 \\ \hline
FLEST & 0.470 & 0.535 & 0.479 & 0.444 & 0.537  & 0.389 & 0.257 & 0.350 \\ \hline
\end{tabular}}
\label{tab:1}
\end{table}
\begin{table}[h!]
\caption{Results of Multi-Client Number Performance}
 \centering
\resizebox{\linewidth}{!}{ \begin{tabular}{|l|llll|llll|}
\hline
                   & \multicolumn{4}{c|}{WN18RR}   & \multicolumn{4}{c|}{FB15k-237} \\ \hline
$\#$ Client &
  \multicolumn{1}{l|}{C = 5} &
  \multicolumn{1}{l|}{C = 10} &
  \multicolumn{1}{l|}{C = 15} &
  C = 20 &
  \multicolumn{1}{l|}{C = 5} &
  \multicolumn{1}{l|}{C = 10} &
  \multicolumn{1}{l|}{C = 15} &
  C = 20 \\ \hline
DistMult (Locally) & 0.070 & 0.052 & 0.055 & 0.037 & 0.123  & 0.078 & 0.071 & 0.067 \\ \hline
ComplEx (Locally)  & 0.004 & 0.004 & 0.003 & 0.002 & 0.120  & 0.074 & 0.066 & 0.056 \\ \hline
RotatE (Locally)   & 0.110 & 0.054 & 0.051 & 0.015 & 0.191  & 0.125 & 0.131 & 0.058 \\ \hline
DistMult (FedE)    & 0.114 & 0.093 & 0.072 & 0.061 & 0.171  & 0.131 & 0.101 & 0.077 \\ \hline
ComplEx (FedE)     & 0.013 & 0.011 & 0.011 & 0.012 & 0.161  & 0.110 & 0.089 & 0.073 \\ \hline
RotatE (FedE)      & \textbf{0.210} & \textbf{0.153} & 0.107 & 0.090 & \textbf{0.261}  & 0.229 & 0.191 & 0.108 \\ \hline
DistMult (FedR)    & 0.121 & 0.109 & 0.091 & 0.089 & 0.168  & 0.100 & 0.087 & 0.081 \\ \hline
ComplEx (FedR)     & 0.019 & 0.015 & 0.015 & 0.013 & 0.170  & 0.116 & 0.105 & 0.092 \\ \hline
RotatE (FedR)      & 0.130 & 0.127 & 0.103 & 0.090 & 0.255  & 0.202 & 0.181 & 0.125 \\ \hline
TuckER (Locally)   & 0.067 & 0.055 & 0.053 & 0.040 & 0.121  & 0.079 & 0.071 & 0.060 \\ \hline
FLEST          & 0.137 & 0.130 & \textbf{0.117} & \textbf{0.093} & 0.252  & \textbf{0.230} & \textbf{0.19}5 & \textbf{0.127} \\ \hline
\end{tabular}}
\label{tab33}
\end{table}
\subsection{Baseline Models}

We compare with the FedE~\cite{chen2021fede} and FedR~\cite{zhang2022efficient} frameworks, in the local training settings. The two federated frameworks can be applied to various baseline models, including DistMult~\cite{yang2015embedding}, ComplEx~\cite{trouillon2016complex}, RotatE~\cite{sun2018rotate} and TuckER~\cite{balazevic2019tucker}. Once the embeddings are provided, the scoring function of our model and other models are displayed in the TABLE~\ref{tab1my}. The scoring function estimates the probability of whether a triplet exists.
FedE introduces a federated mechanism for aggregating entity embeddings. The server maintains a comprehensive table of entity embeddings and their corresponding IDs, enabling it to identify entities for alignment among clients efficiently. In contrast, FedR proposes a federated learning paradigm that emphasizes the aggregation of relation embeddings.

\subsection{Implementation Detail}
For RotatE, DistMult, and ComplEx, we adhere to the same configuration as FedE~\cite{chen2021fede} and FedR~\cite{zhang2022efficient}. All models are trained on a single Nvidia 3090 GPU, with a maximum of 300 communication rounds. Regarding the proposed FLEST, unless otherwise specified, the local update epoch is set to 3, the sparsity factor is 0.5, the Rank is 200, the batch size is 128, and we follow FedR~\cite{zhang2022efficient} for data splitting, where the dataset is eventually divided among each client. We employ the widely-used Adam optimizer to optimize the model update with a learning rate of 0.0005. Additionally, we incorporate a dropout mechanism with a drop rate of 0.3 for parameter regularization and robust training.

\subsection{Single Client Performance}
Initially, in contrast to the federated solutions offered by FedE and FedR, which target different standalone models, our probabilistic tensor decomposition approach brings fundamental modifications to the underlying standalone model. Thus, in this subsection, our main objective is to demonstrate the effectiveness of our standalone solution, aiming to achieve performance \textbf{comparable} to the single machine baselines. We conducted experiments on two datasets, comparing them against several prevalent knowledge graph decomposition models, and reported the results.
Based on the TABLE~\ref{tab:1}, we can observe that the performance of FLEST (with client $\#$ equal to one) is comparable to TuckER, which is also a tensor factorization model, in terms of all metrics on both WN18RR and FB15k-237 datasets. 
This suggests that FLEST performs on par with TuckER when evaluated on a standalone basis.

\subsection{Multi-client Performance}

Subsequently, we evaluated the performance in the federated learning setting involving multiple clients. For this purpose, we carefully partitioned the triplets randomly among the clients without replacement. This random partitioning introduces heterogeneity among all the clients, ensuring a fair and unbiased comparison between different models.
As shown in TABLE~\ref{tab33}, it shows that as the number of clients increases, the overall performance tends to decline. This decline in performance may be attributed to the increased data dispersion caused by a larger number of clients.
Our federated learning approach, FLEST, notably demonstrated exceptional performance compared to the locally trained models. Specifically, FLEST achieved remarkable results regarding Mean Reciprocal Rank (MRR) across the WN18RR and FB15k-237 datasets. 
Although FLEST performed worse than RotatE (FedE) in certain experimental settings, as discussed earlier, FedE methods poses a significant risk of private information leakage. And in other cases, our FLEST model demonstrates the best performance among all baselines.
This significant improvement showcases the effectiveness of FLEST in capturing the underlying patterns and dependencies in the knowledge graph. 
Our FLEST model, within the federated learning framework, emerges as a powerful solution that surpasses the performance of locally trained models. The capability of leveraging distributed knowledge while minimizing the risk of private information leakage proves to be crucial in achieving superior results in knowledge graph completion.

\section{Related Works}
\subsection{Federated Graph Learning}

In general, Federated Graph Learning can be classified into two settings based on the level of structural information~\cite{fu2022federated}. The first setting is Federated Learning (FL) with structured data~\cite{fu2022federated}. In this setting, clients collaborate to train a graph machine-learning model using their local graph data while keeping the graph data decentralized. The second setting is structured Federated Learning (FL)~\cite{fu2022federated}. Structural information exists among the clients in structured FL, forming a client-level graph. 
In the context of this paper, which focuses on knowledge graph partitioning, the main contributions in aspects include FedE~\cite{chen2021fede}, FedR~\cite{zhang2022efficient}, FKGE~\cite{peng2021differentially}. These contributions, including our FLEST, fall under the FL with structured data category.

\subsection{Federated Matrix/Tensor Decomposition}

Federated Matrix/Tensor Factorization is a novel research area applying federated learning techniques to tasks of tensor factorization. It combines federated learning and tensor factorization benefits, enabling collaborative model training across distributed data sources while preserving direct private information. For example, 
TRIP~\cite{kim2017federated} proposes a new federated framework for tensor factorization over horizontally partitioned data.
FedNMF~\cite{si2022federated} is a federated learning approach that effectively tackles the task of federated topic modeling by maximizing the mutual information between input text count features and topic weights. It offers a solution to the challenges associated with federated topic modeling problems.

\section{Conclusion and Discussion}
In this paper, we introduce a tensor factorization based decentralized federated knowledge graph embedding framework that demonstrates excellent performance. In contrast to the potential information leakage of FedE, the decreased performance resulting from the lack of overlapping in FedR, and the issue of model collapse in FKGE, our proposed approach leverages a shared latent dictionary matrix to enable secure and efficient federated information exchange. This innovative FLEST effectively addresses these limitations, ensuring low private information leakage risk and robust performance in the federated learning setting. Our future research focuses on designing an effective federated learning framework for scenarios where different organizations possess diverse types of knowledge graphs requiring personalized embeddings. We also aim to explore KG privacy attack/defense techniques to enhance the privacy and security of federated KGs.

\section{Acknowledgement}
This research was partially supported by APRC - CityU New Research Initiatives (No.9610565, Start-up Grant for New Faculty of City University of Hong Kong), CityU - HKIDS Early Career Research Grant (No.9360163), Hong Kong ITC Innovation and Technology Fund Midstream Research Programme for Universities Project (No.ITS/034/22MS), Hong Kong Environmental and Conservation Fund (No. 88/2022), SIRG - CityU Strategic Interdisciplinary Research Grant (No.7020046, No.7020074), Tencent (CCF-Tencent Open Fund, Tencent Rhino-Bird Focused Research Fund), Huawei (Huawei Innovation Research Program), Ant Group (CCF-Ant Research Fund, Ant Group Research Fund) and Kuaishou.

\bibliography{camera_ready}
\bibliographystyle{IEEEtran}

\vspace{12pt}

\end{document}